\documentclass[conference]{IEEEtran}

\usepackage{cite}
\usepackage{amsmath,amssymb,amsfonts,nccmath}
\usepackage{graphicx}
\usepackage{booktabs}
\usepackage{url}
\usepackage{xcolor}
\usepackage{subcaption}

\title{A Network-Aware Evaluation of Distributed Energy Resource Control in Smart Distribution Systems}

\author{
\IEEEauthorblockN{Houchao Gan}
\IEEEauthorblockA{
Department of Computer Science\\
City University of New York\\
Email: hgan@gradcenter.cuny.edu}
}

\begin{document}
\maketitle

\begin{abstract}
Distribution networks with high penetration of Distributed Energy Resources (DERs) increasingly rely on communication networks to coordinate grid-interactive control. While many distributed control schemes have been proposed, they are often evaluated under idealized communication assumptions, making it difficult to assess their performance under realistic network conditions. This work presents an implementation-driven evaluation of a representative virtual power plant (VPP) dispatch algorithm using a co-simulation framework that couples a linearized distribution-system model with packet-level downlink emulation in ns-3. The study considers a modified IEEE~37-node feeder with high photovoltaic penetration and a primal--dual VPP dispatch that simultaneously targets feeder-head active power tracking and voltage regulation. Communication effects are introduced only on the downlink path carrying dual-variable updates, where per-DER packet delays and a hold-last-value strategy are modeled. Results show that, under ideal communication, the dispatch achieves close tracking of the feeder-head power reference while maintaining voltages within the prescribed limits at selected buses. When realistic downlink delay is introduced, the same controller exhibits large oscillations in feeder-head power and more frequent voltage limit violations. These findings highlight that distributed DER control performance can be strongly influenced by communication behavior and motivate evaluation frameworks that explicitly incorporate network dynamics into the assessment of grid-interactive control schemes. Code available at: \url{https://github.com/Houchao/network-aware-der-control}.
\end{abstract}

\section{Introduction}

Power distribution systems are undergoing significant changes due to the increasing
integration of Distributed Energy Resources (DERs), including photovoltaic generation,
energy storage, and controllable loads.
Unlike traditional centralized generation, DERs are connected at the distribution level
and operate under diverse local conditions.
As DER penetration increases, distribution networks experience higher variability,
bidirectional power flows, and tighter operational constraints, particularly with respect
to voltage regulation and coordination.

Maintaining acceptable voltage profiles in distribution networks with high DER penetration
requires active control strategies.
Distributed DER control has emerged as a practical approach in which control actions are
computed locally or with limited coordination, improving scalability and reducing reliance
on centralized control infrastructure.
However, such distributed control systems rely on communication networks to exchange
coordination information among control entities.
In practical deployments, communication behavior may influence the timing and availability
of control information, which can affect closed-loop control performance.

A substantial body of prior work has focused on the development of DER control and dispatch
algorithms, as well as communication architectures and management platforms to support DER
coordination.
In many of these studies, control performance is evaluated under simplified or abstract
communication assumptions, while communication-focused studies often emphasize network-level
metrics without explicitly examining their impact on power system regulation.
As a result, control algorithms and communication infrastructure are frequently analyzed
as separate components, despite their interaction in deployed systems.

In this work, the interaction between communication behavior and distributed DER control
is evaluated using a co-simulation approach.
A power distribution feeder model is integrated with packet-level communication network
emulation to enable closed-loop evaluation under consistent power system conditions.
The DER control algorithm is kept fixed across all experiments, allowing the effect of
communication behavior on regulation and tracking performance to be observed directly.
The goal of this study is to provide an implementation-driven evaluation of distributed
DER control under realistic communication conditions, rather than to propose new control
or communication algorithms.


\section{Related Work}

Extensive research has been conducted on the control and coordination of DERs in power distribution systems.
Prior work has proposed a variety of DER control and dispatch strategies aimed at enabling
scalable and flexible operation under high DER penetration.
These include virtual power plant (VPP) formulations and distributed coordination mechanisms,
which seek to aggregate heterogeneous DERs and provide grid services while reducing reliance
on centralized control architectures \cite{EmilianoVPP2018,ZhouDisVPP2019}.

In addition to control algorithms, communication architectures supporting DER applications
have been studied to enable information exchange among geographically distributed assets.
These efforts address practical concerns such as coordination, data aggregation, and
interoperability across heterogeneous communication technologies and system interfaces
\cite{DisCommVPP2018,VPP18,CommAna2019}.
At a higher level, Distributed Energy Resource Management Systems (DERMS) have been developed
to provide supervisory monitoring and control of large DER fleets and to facilitate their
integration into utility operations \cite{epriDERMS2018,PGEderms2019}.

Most existing studies on DER control evaluate algorithmic performance under abstract or
idealized communication assumptions, where coordination information is implicitly assumed
to be available when needed.
For example, voltage regulation strategies based on local feedback or simplified coordination
models are commonly analyzed without explicitly modeling communication behavior, and the
limitations of purely local control in the absence of communication have been highlighted
in prior work \cite{Bolognani2019TCNS}.
More recent studies have begun to incorporate imperfect, asynchronous, or delayed communication
models into distributed voltage control formulations, indicating that non-ideal communication
can influence regulation performance \cite{Wang2023TSG,Magnusson2020TSG}.
However, these effects are often examined within specific analytical models rather than through
implementation-driven evaluation. Studies focused on DER communication architectures typically emphasize network-level
performance metrics such as latency, throughput, or reliability, without explicitly evaluating
their impact on closed-loop power system behavior.
As a result, the combined effect of practical communication behavior on DER regulation and
tracking performance is not always directly evaluated.

This work is situated within this body of work by evaluating distributed DER control
using a co-simulation environment that explicitly models both the power distribution system
and the communication network.
Rather than proposing new control or communication algorithms, the focus is placed on examining
how communication behavior influences the observed performance of an existing DER control
scheme under consistent power system conditions in real-time.


\section{Problem Statement and Recap of VPP Dispatch}

A Virtual Power Plant (VPP) dispatch process is commonly used to coordinate a collection of
Distributed Energy Resources (DERs) connected to a distribution feeder.
At each control interval, measurements (such as bus voltage magnitudes and feeder-head
power) are collected, and setpoints are computed for DER inverters.
In this work, the VPP dispatch mechanism is treated as a fixed, known control policy, and
the focus is placed on evaluating how communication behavior affects its closed-loop
voltage regulation and feeder-head tracking performance.

A distribution feeder with $N+1$ nodes is considered, with nodes indexed by
$\mathcal{N}\cup\{0\}$, where node $0$ denotes the secondary of the distribution transformer
or feeder head.
Let $S_i = P_i + jQ_i$ denote the complex power injected at node $i\in\mathcal{N}$ and
let $P_{l,i},Q_{l,i}$ denote the active and reactive loads at node $i$.
The net complex power injection vector is defined as
\begin{equation}
\mathbf{s}_{inj} := [S_1,\dots,S_N]^T \in \mathbb{C}^N,
\end{equation}
where
\begin{equation}
S_i =
\begin{cases}
P_i - P_{l,i} + j(Q_i - Q_{l,i}), & i \in \mathcal{G}, \\
- P_{l,i} - jQ_{l,i}, & i \in \mathcal{N}\setminus\mathcal{G},
\end{cases}
\end{equation}
so that DER injections and loads are combined into net nodal injections.

Let $\mathbf{p}_{inj} := \Re\{\mathbf{s}_{inj}\}$ and
$\mathbf{q}_{inj} := \Im\{\mathbf{s}_{inj}\}$ denote the active and reactive net injection
vectors.
By linearizing the AC power-flow equations around a nominal operating point
$\mathbf{v}_{nom}$, a LinDistFlow-type approximation is obtained for the nodal voltage
magnitudes and feeder-head power:
\begin{equation}
\begin{aligned}
|\mathbf{v}| &\approx \mathbf{A}\,\mathbf{p}_{inj} + \mathbf{B}\,\mathbf{q}_{inj} + \mathbf{c}, \\
\begin{bmatrix} P_0 \\ Q_0 \end{bmatrix}
&\approx \mathbf{M}\,\mathbf{p}_{inj} + \mathbf{N}\,\mathbf{q}_{inj} + \mathbf{o},
\end{aligned}
\label{eq:lindistflow_model}
\end{equation}
where $|\mathbf{v}| \in \mathbb{R}^N$ denotes the vector of nodal voltage magnitudes and
$P_0,Q_0$ denote the active and reactive powers at the feeder head.
The matrices $\mathbf{A},\mathbf{B}\in\mathbb{R}^{N\times N}$,
$\mathbf{M},\mathbf{N}\in\mathbb{R}^{2\times N}$ and the vectors
$\mathbf{c}\in\mathbb{R}^N$, $\mathbf{o}\in\mathbb{R}^2$ are obtained by linearizing the
AC power-flow equations around the nominal operating point, following the approach in
\cite{EmilianoVPP2018}.
In the subsequent VPP dispatch formulation, the voltage magnitudes $|\mathbf{v}|$ and
feeder-head active power $P_0$ are evaluated using \eqref{eq:lindistflow_model} and used
in the voltage regulation and tracking objectives.

The real-time optimal VPP dispatch with voltage regulation is formulated using a
regularized Lagrangian and primal--dual gradient iterations.
Let $[V^{\min},V^{\max}]$ denote the acceptable voltage range at the monitored buses $\mathcal{M}$.
Define dual variables $\boldsymbol{\gamma}^{t_k}=[\gamma_1^{t_k},\dots,\gamma_M^{t_k}]^T$
and $\boldsymbol{\mu}^{t_k}=[\mu_1^{t_k},\dots,\mu_M^{t_k}]^T$ associated with the lower
and upper voltage bounds at time $t_k$, respectively.
Let $\lambda^{t_k}$ and $\zeta^{t_k}$ be the dual variables associated with feeder-head
power tracking constraints with tolerance $E^{t_k}>0$.
Collect all dual variables in
$\mathbf{d}^{t_k}:=\{\boldsymbol{\gamma}^{t_k},\boldsymbol{\mu}^{t_k},\lambda^{t_k},\zeta^{t_k}\}$.
With these definitions, the regularized Lagrangian at time $t_k$ is given by
\begin{equation}
\begin{aligned}
\mathcal{L}^{t_k}(\mathbf{p},\mathbf{q},\mathbf{d}) :=\;&
\sum_{i\in\mathcal{G}} f_i^{t_k}(P_i,Q_i) \\
&+ \sum_{n\in\mathcal{M}}
\gamma_n^{t_k}\big(V^{\min}-|V_n^{t_k}(\mathbf{p},\mathbf{q})|\big) \\
&+ \sum_{n\in\mathcal{M}}
\mu_n^{t_k}\big(|V_n^{t_k}(\mathbf{p},\mathbf{q})|-V^{\max}\big) \\
&+ \lambda^{t_k}\big(P_0^{t_k}(\mathbf{p},\mathbf{q})-P_{0,\mathrm{set}}^{t_k}-E^{t_k}\big) \\
&+ \zeta^{t_k}\big(P_{0,\mathrm{set}}^{t_k}-P_0^{t_k}(\mathbf{p},\mathbf{q})-E^{t_k}\big) \\
&+ \frac{\nu}{2}\sum_{i\in\mathcal{G}}\big[(P_i^{t_k})^2+(Q_i^{t_k})^2\big]
- \frac{\epsilon}{2}\|\mathbf{d}^{t_k}\|_2^2,
\end{aligned}
\label{eq:vpp_lagrangian}
\end{equation}
where $\nu>0$ and $\epsilon>0$ are regularization coefficients.
The primal variables are stacked as
$\mathbf{p}=[P_1^{t_k},\dots,P_{|\mathcal{G}|}^{t_k}]^T$ and
$\mathbf{q}=[Q_1^{t_k},\dots,Q_{|\mathcal{G}||}^{t_k}]^T$, and
$P_{0,\mathrm{set}}^{t_k}$ denotes the desired feeder-head active power at time $t_k$.

The local objectives $f_i^{t_k}(P_i,Q_i)$ capture DER- and utility-level preferences.
For PV inverters, a quadratic objective is adopted to encourage utilization of available
active power and limit reactive power effort:
\begin{equation}
f_i^{t_k}(P_i,Q_i)
= c_p\big(P_{av,i}^{t_k}-P_i^{t_k}\big)^2
+ c_q\big(Q_i^{t_k}\big)^2,
\label{eq:pv_objective}
\end{equation}
where $P_{av,i}^{t_k}$ denotes the available active power at inverter $i$, and $c_p>0$ and
$c_q>0$ are weighting coefficients.
The feasible operating region for each inverter is captured by the capability set
\begin{equation}
\mathcal{Y}_i^{t_k}
=\big\{(P_i,Q_i): 0\le P_i\le P_{av,i}^{t_k},\; P_i^2+Q_i^2\le S_i^2\big\},
\label{eq:inverter_feasible}
\end{equation}
where $S_i$ is the apparent power rating of inverter $i$.

At each discrete time instant $t_k$, the real-time VPP dispatch problem is implemented via
a primal--dual procedure that seeks a saddle point of the Lagrangian:
\begin{equation}
\max_{\mathbf{d}^{t_k}\in\mathbb{R}_+^{2M+2}}
\min_{(P_i,Q_i)\in\mathcal{Y}^{t_k}}
\mathcal{L}^{t_k}(\mathbf{p},\mathbf{q},\mathbf{d}^{t_k}),
\label{eq:optialgo_compact}
\end{equation}
where $\mathcal{Y}^{t_k}:=\prod_{i\in\mathcal{G}}\mathcal{Y}_i^{t_k}$.
Primal and dual variables are updated using gradient steps with stepsize $\alpha>0$ and
dual regularization coefficient $\epsilon>0$.
The detailed update equations are not repeated here; instead, the resulting mapping from
measurements to DER setpoints is treated as a fixed VPP dispatch controller.

The formulation above defines the control variables, constraints, and performance
objectives considered in this work.
Dispatch setpoints are computed based on feeder measurements and applied to DER inverters
in a discrete-time closed-loop manner.
Voltage regulation and feeder-head tracking performance are evaluated using these
definitions, while the effect of communication behavior on the timing and availability of
dual variables and control inputs is examined in the subsequent sections.

\section{Communication Model and Distributed VPP Dispatch with Dual Variables}
\label{sec:comm_model}

The distributed VPP dispatch considered in this work follows a feeder--DER coordination
architecture based on a primal--dual structure.
At each discrete control instant $t_k$, voltage magnitudes $|\widehat{V}_n^{t_k}|$ at monitored buses
$\mathcal{M}$ and feeder-head power are evaluated, and the dual variables
$\boldsymbol{\gamma}^{t_k}$, $\boldsymbol{\mu}^{t_k}$, $\lambda^{t_k}$, and $\zeta^{t_k}$
associated with the voltage and tracking constraints in \eqref{eq:vpp_lagrangian} are
updated at a feeder-level controller.

In this implementation, the components $\boldsymbol{\gamma}^{t_k}$ and $\boldsymbol{\mu}^{t_k}$ encode the
marginal penalties on the lower and upper voltage limits, while $\lambda^{t_k}$ and
$\zeta^{t_k}$ encode the penalties associated with feeder-head power tracking; together,
these dual variables act as coordination signals for DER-level decisions through the
update \eqref{eq:local_der_update}.

Upon receiving the feeder-level dual variables, each DER performs a projected
gradient step on the Lagrangian~\eqref{eq:vpp_lagrangian} to update its local
active and reactive power setpoints.
For DER $i\in\mathcal{G}$, the update takes the form

\begin{equation}
\begin{aligned}
\begin{bmatrix}
P_i^{t_{k+1}} \\[2pt]
Q_i^{t_{k+1}}
\end{bmatrix}
&=
\operatorname{proj}_{\mathcal{Y}_i^{t_k}}
\Bigg\{
\begin{bmatrix}
P_i^{t_k} \\[2pt]
Q_i^{t_k}
\end{bmatrix}
- \alpha\,
\nabla_{[P_i,Q_i]}\mathcal{L}^{t_k}(\mathbf{p},\mathbf{q},\mathbf{d})
\\[-2pt]
&\hspace{3.2em}
\Big|_{P_i^{t_k},\,Q_i^{t_k},\,\mathbf{d}^{t_{k+1}}}
\Bigg\}
\end{aligned}
\label{eq:local_der_update}
\end{equation}

where $\operatorname{proj}_{\mathcal{Y}_i^{t_k}}(\cdot)$ denotes the projection onto
the inverter capability set $\mathcal{Y}_i^{t_k}$ defined in~\eqref{eq:inverter_feasible},
$\alpha>0$ is a stepsize, and $\mathbf{d}^{t_{k+1}}$ collects the updated dual variables
(e.g., $\boldsymbol{\gamma}^{t_{k+1}}, \boldsymbol{\mu}^{t_{k+1}}, \lambda^{t_{k+1}},
\zeta^{t_{k+1}}$).

Let
\[
\mathbf{d}^{t_k}
:= \{\boldsymbol{\gamma}^{t_k},\,\boldsymbol{\mu}^{t_k},\,\lambda^{t_k},\,\zeta^{t_k}\}
\]
denote the collection of current dual variables that exist on each DER.
In the ideal-communication case, the feeder-level controller updates
$\mathbf{d}^{t_{k+1}}$ at each control step and broadcasts it to all DER
controllers, which then each DER apply the primal update~\eqref{eq:local_der_update} using
$\mathbf{d}^{t_{k+1}}$.

Due to communication effects such as delay or packet loss on the downlink,
a DER may not receive the most recent dual-variable packet within a control
interval.
Let $\tilde{\mathbf{d}}^{t_k}$ denote the dual variables actually available at
a DER at time $t_k$.
A hold-last-value recovery strategy is adopted, modeled as
\begin{equation}
\tilde{\mathbf{d}}^{t_k} =
\begin{cases}
\mathbf{d}^{t_{k+1}}, & \text{if a new dual packet is received,} \\[4pt]
\mathbf{d}^{t_k}, & \text{otherwise,}
\end{cases}
\label{eq:dual_delay}
\end{equation}
so that, in the absence of a new packet, the most recently received dual
variable is reused.
In the delayed case, the local update has the same form as
\eqref{eq:local_der_update}, with $\mathbf{d}^{t_{k+1}}$ replaced by
$\mathbf{d}^{t_{k}}$ in the gradient evaluation.

As illustrated in Fig.~\ref{fig:comm_arch}, the downlink communication
behavior directly influences the closed-loop voltage regulation performance
of the distributed VPP dispatch, even though the underlying control
formulation at both the feeder and DER levels remains unchanged.

\begin{figure}[t]
  \centering
  \includegraphics[width=\columnwidth]{figures/communcation_model.png}
  \caption{Communication architecture for the distributed VPP dispatch.
  A feeder-level controller (DERMS) located at the feeder head exchanges
  information with grid-edge devices over a neighborhood area network.
  Voltage measurements are collected from DER locations via uplink channels,
  while coordination signals (dual variables and dispatch commands) are
  transmitted to DER controllers via downlink channels.
  Within each building, a home area network connects local DER inverters and
  applications to the grid-edge device.
  In the experiments, only the downlink path from the feeder-level controller
  to DER controllers is modeled in ns-3; uplink measurement delivery is
  assumed to be available and is not delay-modeled.}
  \label{fig:comm_arch}
\end{figure}


\section{Experimental Setup}

\begin{figure}[t]
  \centering
  \includegraphics[width=\columnwidth]{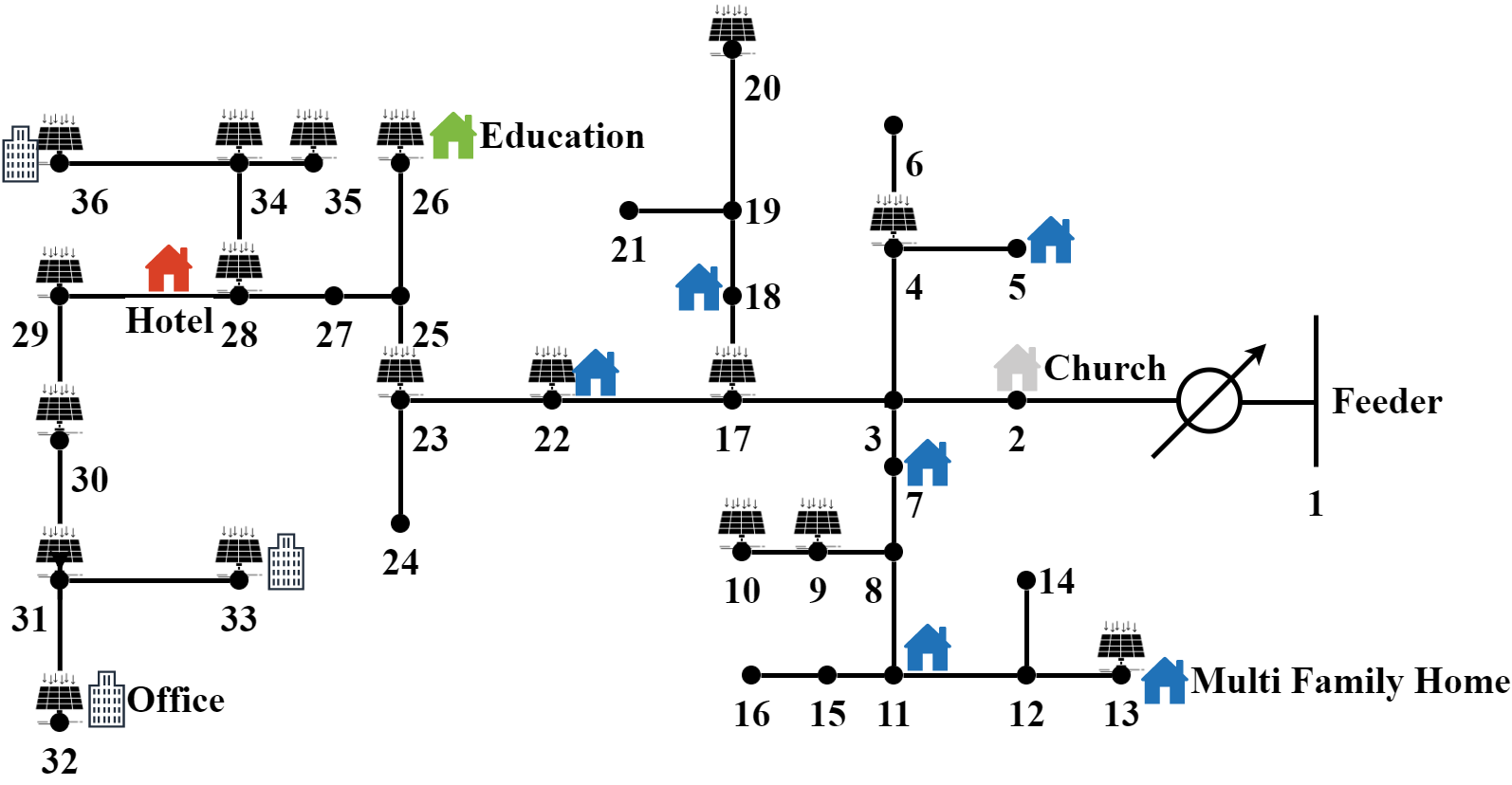}
  \caption{Modified IEEE 37-node test feeder used in this work. PV systems are placed at
  nodes 4, 9, 10, 13, 17, 20, 22, 23, 26, 28, 29, 30, 32, 33, 34, 35, and 36.}
  \label{fig:ieee37_feeder}
\end{figure}

The experimental evaluation is conducted on a modified IEEE 37-node test feeder, shown in
Fig.~\ref{fig:ieee37_feeder}.
Rooftop photovoltaic (PV) systems are deployed at eighteen buses along the feeder, located
at nodes 4, 9, 10, 13, 17, 20, 22, 23, 26, 28, 29, 30, 32, 33, 34, 35, and 36.
Each PV system is modeled as an inverter-interfaced DER with an installed capacity of
100~kW and the capability to provide both active and reactive power support subject to
inverter apparent power constraints.

\begin{figure}[t]
  \centering
  \begin{subfigure}{0.76\columnwidth}
    \centering
    \includegraphics[width=\linewidth]{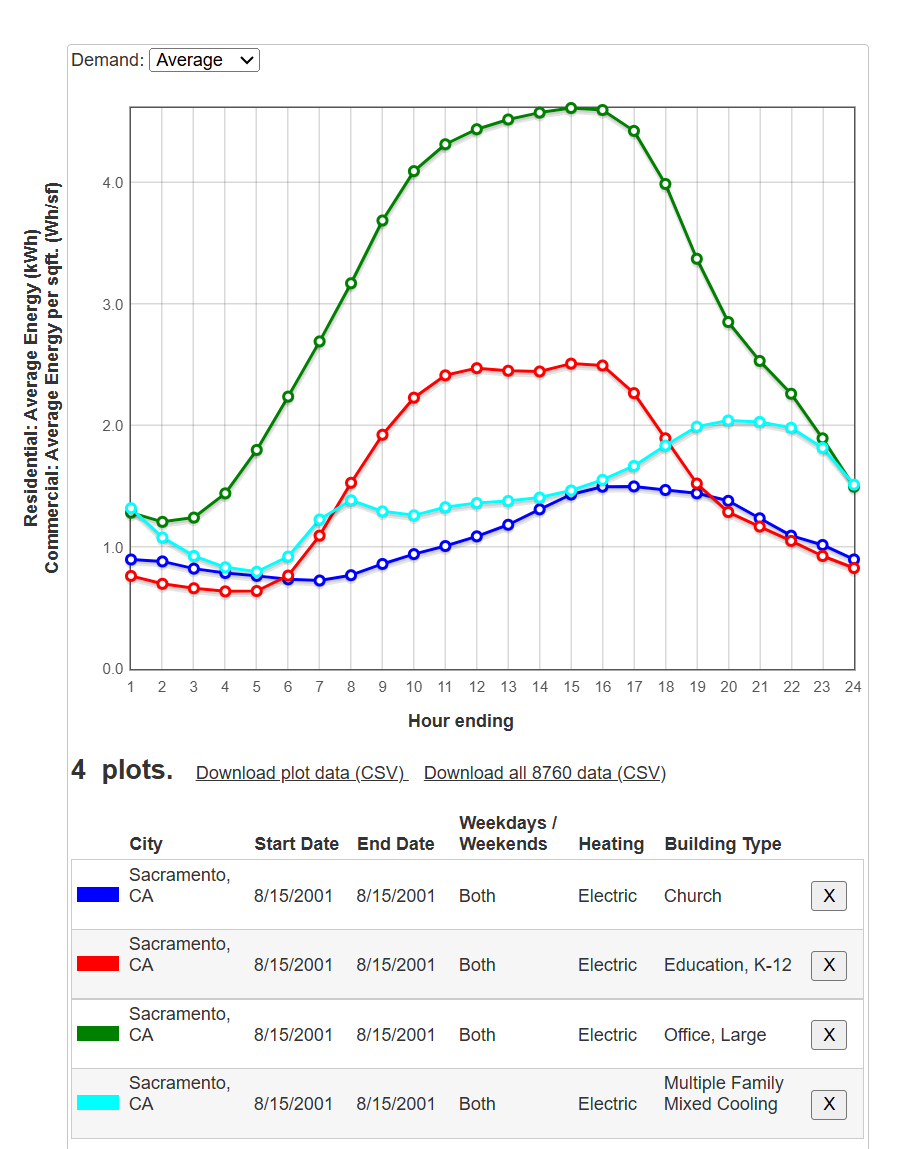}
    \caption{EPRI load profiles at selected Phase-C buses.}
    \label{fig:profiles_load}
  \end{subfigure}

  \vspace{0.3em}

  \begin{subfigure}{0.76\columnwidth}
    \centering
    \includegraphics[width=\linewidth]{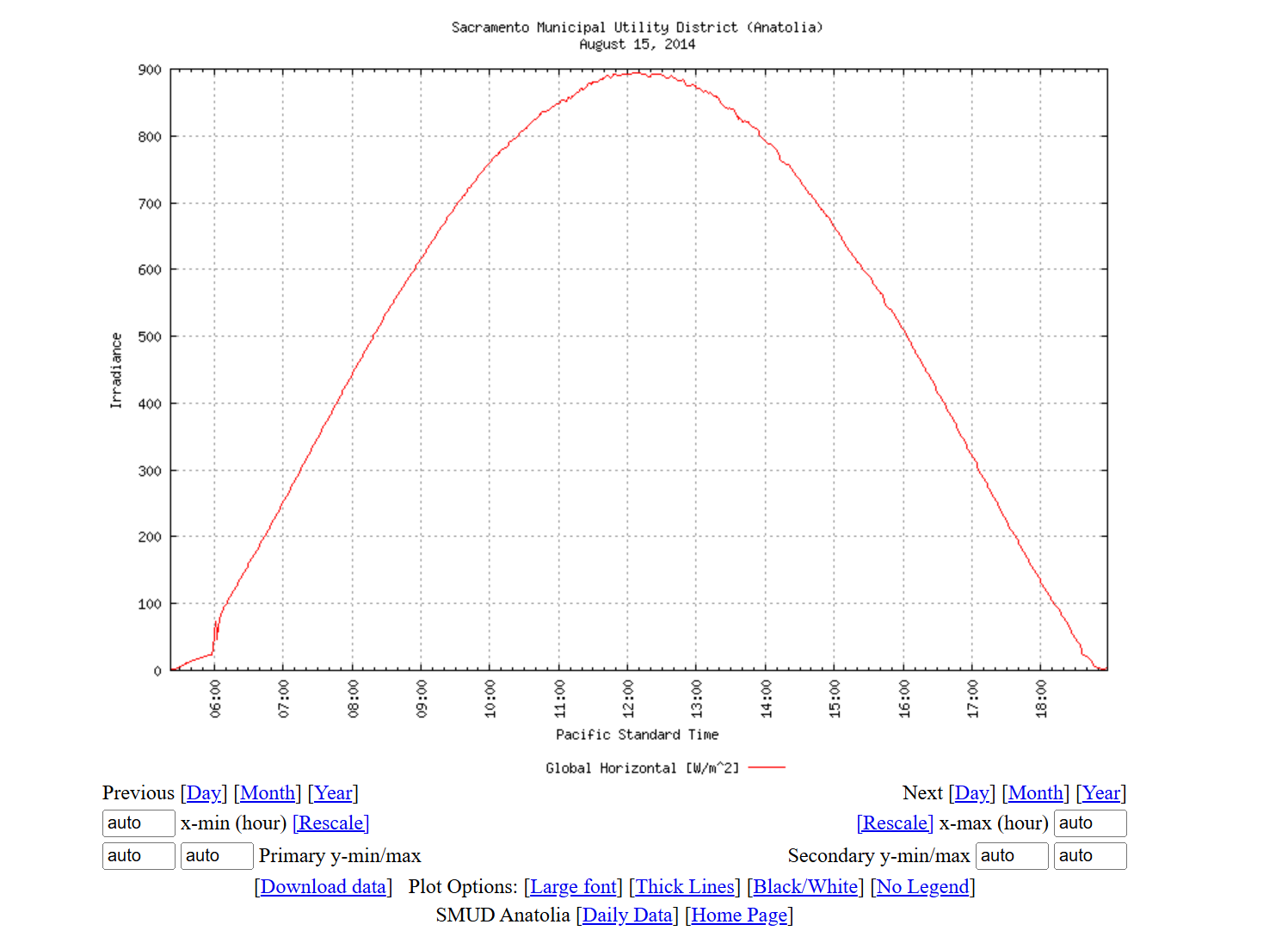}
    \caption{PV generation profiles derived from NREL MIDC irradiance.}
    \label{fig:profiles_pv}
  \end{subfigure}

  \caption{Representative 1-hour-resolution time-series profiles used in the experiments:
  (a) EPRI load profiles applied to selected Phase-C buses in Sacramento, CA; and
  (b) PV generation profiles obtained from August~15 NREL MIDC solar irradiance data,
  processed and scaled to the installed PV capacity.}
  \label{fig:profiles}
\end{figure}

PV generation profiles are derived from solar irradiance measurements collected in
Sacramento, California, obtained from the National Renewable Energy Laboratory (NREL)
Measurement and Instrumentation Data Center (MIDC).
Due to the lack of temporally matched load and solar datasets, irradiance data from
August~15,~2004 are used, while feeder load profiles correspond to August~15,~2001.
Although the datasets originate from different years, both correspond to the same season
and similar operating conditions, preserving seasonal consistency between load demand and
solar availability.
The original irradiance data are provided at an hourly resolution and are converted to a
one-second resolution using linear interpolation to match the control update interval;
representative profiles are shown in Fig.~\ref{fig:profiles}.

Voltage regulation limits are set as $V^{min}=0.95$~p.u.\ and $V^{max}=1.05$~p.u.
Control parameters are set as $\nu=10^{-3}$, $\epsilon=10^{-4}$, $E^{t_k}=0.001$, and
$\alpha=0.1$.
The local PV objective used in the DER update \eqref{eq:local_der_update} is selected as
\begin{equation}
f_i^{t_k}(P_i,Q_i)=c_p\!\left(P_{av,i}^{t_k}-P_i^{t_k}\right)^{2}+c_q\!\left(Q_i^{t_k}\right)^{2},
\label{eq:pv_objective}
\end{equation}
where $c_p=3$, $c_q=1$, and $i\in\mathcal{G}$.
Feeder-level active power setpoints $P_{0,\mathrm{set}}^{t_k}$ are applied over the interval
from 12:00 to 14:00.

The distributed VPP dispatch operates in discrete time with a fixed control interval of
one second.
At each control step, voltage magnitudes are measured at DER locations and are transmitted
through the communication network to the feeder-level controller.
Based on these voltage measurements, dual variables associated with voltage regulation
constraints are updated at the feeder and transmitted to DER controllers.
Each DER computes local active and reactive power setpoints using the received dual
variables and local inverter constraints, and applies these setpoints until the next
control update.

\subsection{ns-3 Downlink Simulation}

The downlink communication path from the feeder-level controller to DER controllers is
emulated using the ns-3 discrete-event network simulator.
The objective of this emulation is to obtain packet-level delay traces for control
packets carrying dual-variable updates, under a simple but explicit model of the
neighborhood area network.

A star topology is used, with one controller node representing the feeder-level
controller (DERMS) and $N_{\mathrm{DER}}=18$ DER nodes representing the DER controllers.
Each DER node is connected to the controller by an independent point-to-point link with
a data rate of 10~Mb/s and a fixed propagation delay of $\texttt{linkDelayMs}$ milliseconds
(default 1~ms in the simulations).
No background traffic is injected, so all queuing and delay originate from propagation
delay and the additional random jitter described below.

The Internet stack is installed on all nodes and UDP is used as the transport protocol.
For each DER $i\in\{1,\dots,N_{\mathrm{DER}}\}$, a UDP server is created on the DER node
and a UDP client is created on the controller node targeting that server.
The controller sends one UDP packet per DER every control interval with period
$\texttt{sendInterval}=1$~s.
Each packet corresponds to one downlink dual-variable update for DER $i$ and has a payload
of 64~bytes.
The total simulation time is set to $\texttt{simTime}=86400$~s, corresponding to 24~hours
of 1~s control intervals.

To measure end-to-end downlink delay, each packet is time-stamped at the controller just
before transmission.
Specifically, a \texttt{TimestampTag} is attached to the packet using the current
simulation time $\mathrm{Now()}$ in a transmit trace callback.
At the DER side, upon reception of a packet by the UDP server, an additional per-DER
random jitter is applied and the resulting arrival time is logged.
For each DER $i$, an independent uniform random variable
$J_i\sim\mathcal{U}(\texttt{jitterMinMs},\texttt{jitterMaxMs})$ is created, with
$\texttt{jitterMinMs}=1$~ms and $\texttt{jitterMaxMs}=150$~ms in the experiments.
When a packet for DER $i$ is received, a delayed logging event is scheduled
$J_i$ seconds after the reception time.
At the scheduled time, the original transmit timestamp is read from the packet tag and the
end-to-end delay is computed as the difference between the delayed reception time and the
transmit time.
This delay is expressed in milliseconds and written to a CSV log file together with the
DER index and the reception time (in seconds).

The resulting CSV file \texttt{der\_downlink\_delay.csv} contains one line per successfully
received packet with the format
\[
\texttt{rx\_time\_sec},\ \texttt{der\_id},\ \texttt{delay\_ms}.
\]
These traces are post-processed to derive, for each DER and each 1~s control interval,
the effective downlink delay in units of control steps.
If at least one packet for DER $i$ is received within a given control interval, the most
recent packet in that interval is treated as delivering the updated dual variables with
a delay corresponding to the measured end-to-end delay.
If no packet is received for DER $i$ during a control interval, the last successfully
received dual-variable values are reused, consistent with the hold-last-value recovery
strategy in \eqref{eq:dual_delay}.

By constructing downlink delay sequences in this way and applying them to the dual
variables used in the distributed VPP dispatch, the impact of packet-level communication
behavior on voltage regulation and feeder-head tracking can be evaluated under otherwise
identical power system conditions.


\section{Results}

The numerical experiments evaluate the distributed VPP dispatch on the modified
IEEE~37-node feeder over the two-hour window from 12:00 to 14:00.
Two communication conditions are compared.
In the \emph{ideal-communication} case, dual-variable updates are assumed to be delivered
to all DER controllers within each one-second control interval without delay.
In the \emph{downlink-delay} case, dual-variable updates are transmitted through the ns-3
network, which introduces packet-level delay and implements the hold-last-value strategy
described in Section~\ref{sec:comm_model}.

\begin{figure*}[t]
    \centering
    \begin{subfigure}{0.48\textwidth}
        \centering
        \includegraphics[width=\linewidth]{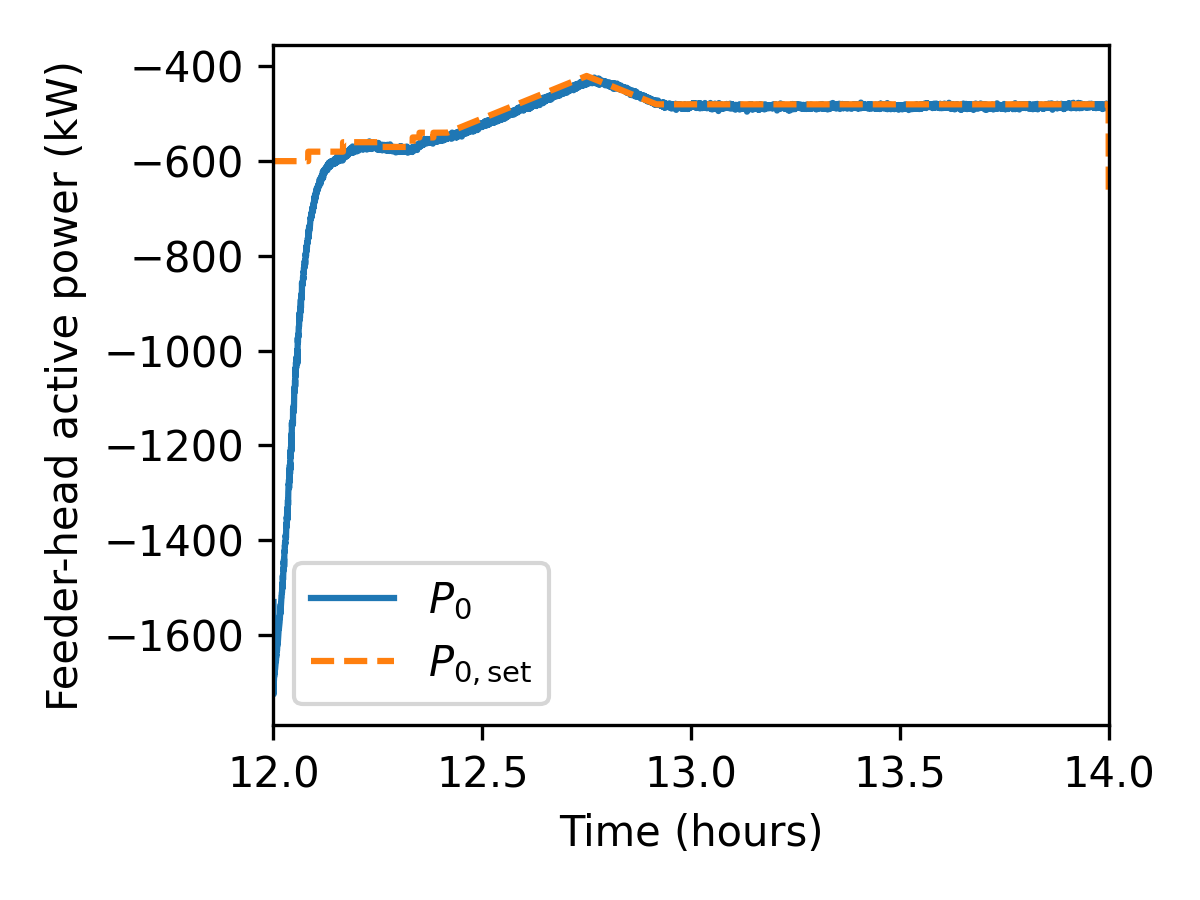}
        \caption{Feeder-head active power $P_0$ and reference $P_{0,\mathrm{set}}$ under ideal communication.}
        \label{fig:ideal_P0}
    \end{subfigure}\hfill
    \begin{subfigure}{0.48\textwidth}
        \centering
        \includegraphics[width=\linewidth]{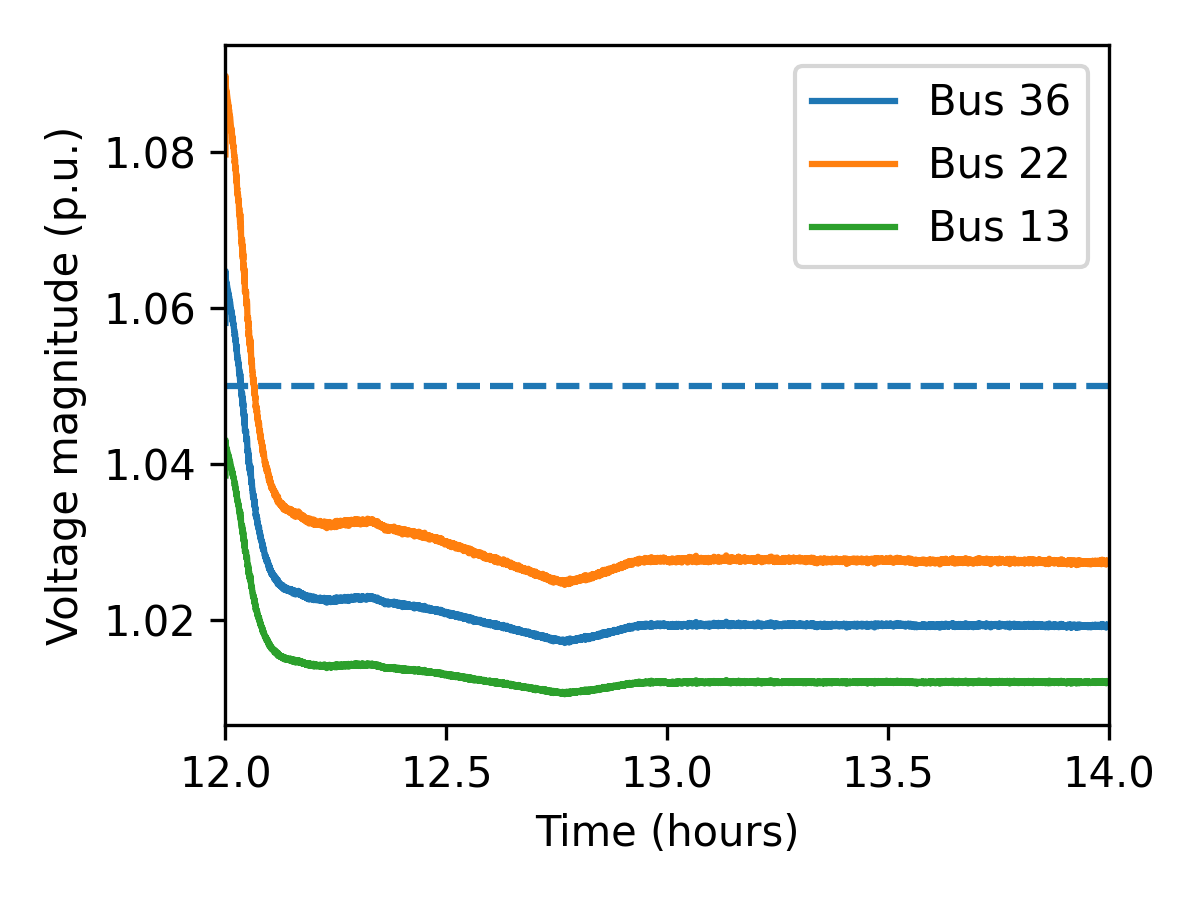}
        \caption{Voltage magnitudes at buses 36, 22, and 13 under ideal communication.}
        \label{fig:ideal_V}
    \end{subfigure}
    \caption{Closed-loop behavior of the distributed VPP dispatch with ideal communication.}
    \label{fig:ideal_results}
\end{figure*}

Figure~\ref{fig:ideal_results} shows the closed-loop behavior under ideal communication.
In Fig.~\ref{fig:ideal_results}(a), the feeder-head active power $P_{0}$ tracks the
reference trajectory $P_{0,\mathrm{set}}$ after a short transient at the beginning of the
window.
The tracking signal remains close to the reference throughout 12:00--14:00, and no large,
systematic offsets are observed.
Figure~\ref{fig:ideal_results}(b) reports the voltage magnitudes at buses~36, 22, and~13.
After the initial transient, voltages settle into a narrow band around $1.0$~p.u.
The upper voltage limit at $1.05$~p.u.\ is rarely reached, and no significant oscillatory
behavior is visible.
These plots indicate that, in the absence of communication delay, the VPP dispatch is able
to achieve both feeder-head power tracking and acceptable voltage regulation at the
selected buses.

\begin{figure*}[t]
    \centering
    \begin{subfigure}{0.48\textwidth}
        \centering
        \includegraphics[width=\linewidth]{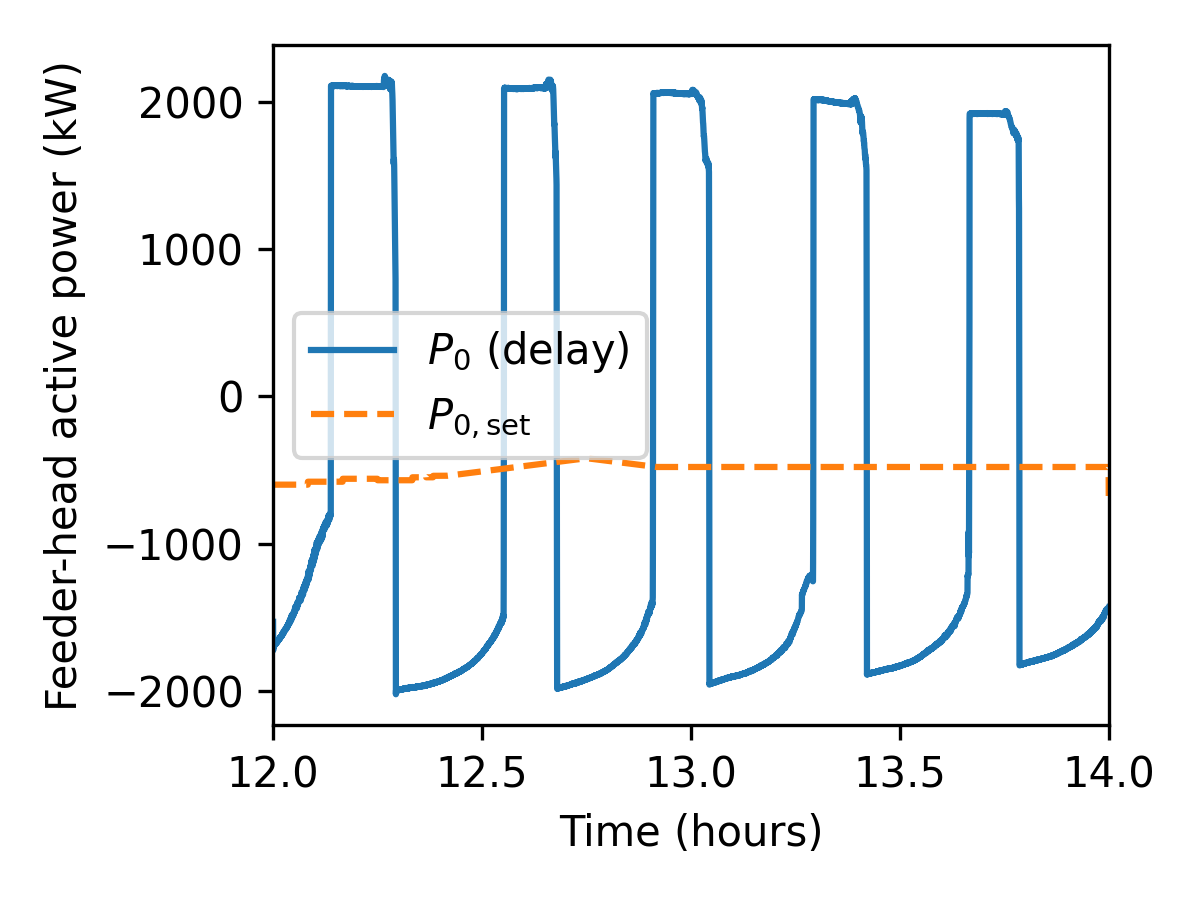}
        \caption{Feeder-head active power $P_0$ and reference $P_{0,\mathrm{set}}$ with downlink delay.}
        \label{fig:delay_P0}
    \end{subfigure}\hfill
    \begin{subfigure}{0.48\textwidth}
        \centering
        \includegraphics[width=\linewidth]{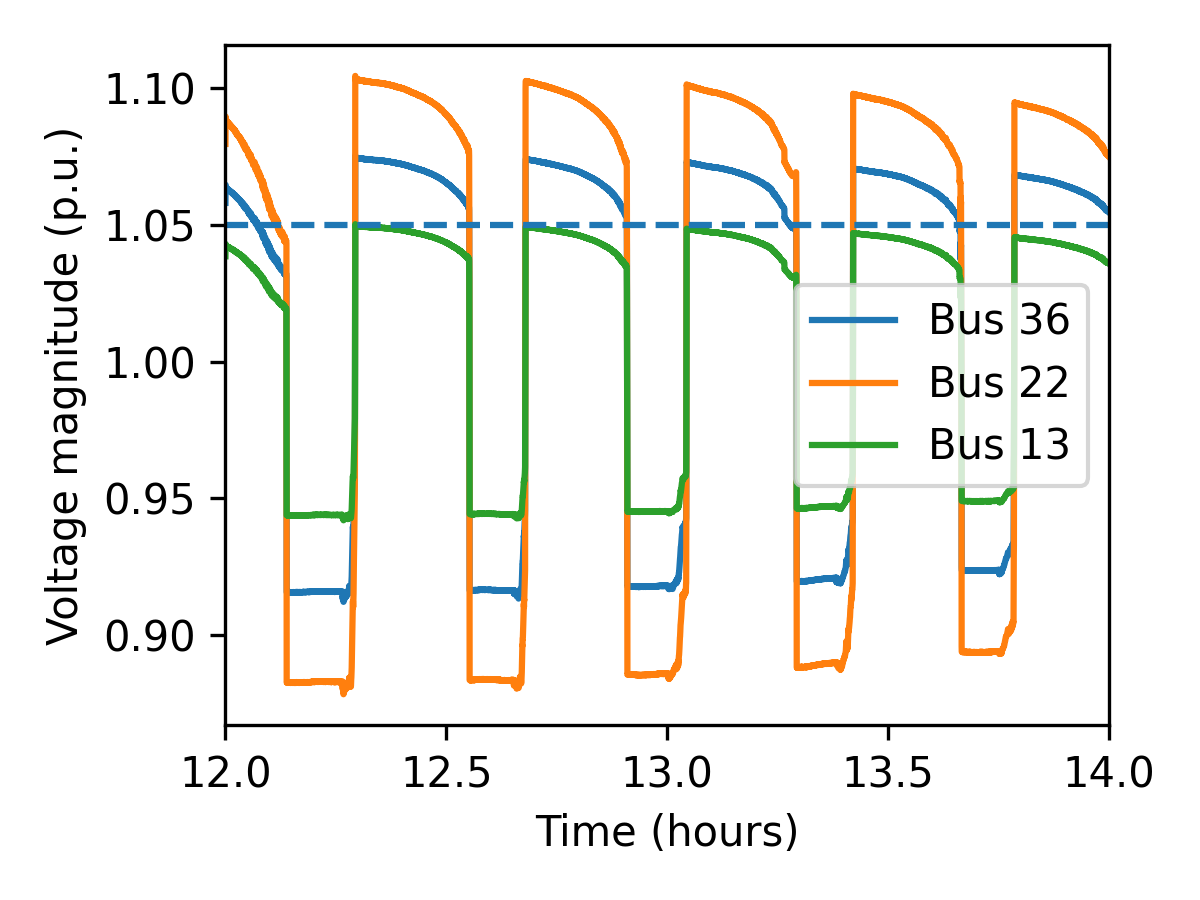}
        \caption{Voltage magnitudes at buses 36, 22, and 13 with downlink delay.}
        \label{fig:delay_V}
    \end{subfigure}
    \caption{Closed-loop behavior of the distributed VPP dispatch with ns-3 downlink delay and hold-last-value strategy.}
    \label{fig:delay_results}
\end{figure*}

Figure~\ref{fig:delay_results} presents the corresponding results when downlink delay is
introduced.
In Fig.~\ref{fig:delay_results}(a), the feeder-head active power $P_{0}$ no longer follows
$P_{0,\mathrm{set}}$ smoothly.
Instead, large oscillations appear, with repeated overshoots and undershoots relative to
the slowly varying reference.
The oscillatory pattern is aligned with the timing of the dual-variable updates, reflecting
the fact that delayed coordination signals cause the dispatch to react to outdated
information.

The impact on voltage regulation is illustrated in
Fig.~\ref{fig:delay_results}(b), which again shows the voltage magnitudes at buses~36, 22,
and~13.
Compared with the ideal-communication case, the voltages exhibit more frequent and larger
excursions.
The trajectories repeatedly rise above the $1.05$~p.u.\ limit and drop closer to the lower
end of the acceptable range.
The shape of the trajectories follows the oscillations in $P_{0}$, indicating that the
voltage profile is strongly influenced by the delayed feeder-head tracking.
Although the underlying control formulation is unchanged, the introduction of realistic
downlink delay leads to visibly degraded performance in both feeder-head tracking and
voltage regulation.

Overall, the two sets of plots highlight that the distributed VPP dispatch is sensitive to
downlink communication dynamics.
When the same controller is evaluated under identical power-system conditions but with
different communication assumptions, the observed closed-loop behavior changes
substantially.

\section{Limitations and Future Work}

This study is deliberately narrow in scope.
The distribution grid is represented by a linearized model of a single modified
IEEE~37-node feeder with inverter-based PV only, and the same primal--dual VPP dispatch
is used in all cases.
Nonlinear AC effects, other device types, and alternative controllers are not considered.
On the communication side, only downlink delivery of dual variables is modeled in ns-3,
using a simple star topology with independent point-to-point links and uniformly
distributed jitter; uplink measurements are assumed to be reliable and delay-free.

Future work will extend the framework along three directions.
First, more detailed power-system models and heterogeneous DER portfolios will be
incorporated to test scalability and robustness.
Second, uplink delay, shared-media contention, and background traffic will be added to the
communication model to represent neighborhood-area networks more realistically.
Third, alternative delay-aware or event-triggered control schemes will be implemented and
compared against the baseline dispatch, enabling joint assessment of control design and
communication configuration for grid-interactive DER operation.

\section{Conclusion}

This work has presented a co-simulation study of distributed VPP dispatch that couples a linearized distribution-system model, a representative primal--dual controller, and an ns-3-based model of the downlink communication network on a modified IEEE~37-node feeder with high PV penetration. The results show that, under ideal communication, the dispatch closely tracks the feeder-head power reference while keeping voltages near the desired band, whereas introducing packet-level downlink delay with a hold-last-value strategy leads to pronounced oscillations in feeder-head power and more frequent voltage limit excursions at selected buses. These observations indicate that evaluating DER control algorithms without explicitly modeling communication behavior can overestimate achievable performance and motivate future work on extending the framework to include uplink effects, more realistic network configurations, and delay-aware or event-triggered control designs for robust grid-interactive DER operation.


\bibliographystyle{IEEEtran}
\bibliography{references}
\end{document}